\documentclass[journal]{IEEEtran}
\usepackage{color}

\usepackage{algorithm}
\usepackage{algorithmic}
\usepackage{array}
\usepackage{hyperref}       
\usepackage{url}            
\usepackage{booktabs}       
\usepackage{amsfonts}       
\usepackage{nicefrac}       
\usepackage{microtype}      
\usepackage{cite}
\usepackage[tableposition=top]{caption}
\usepackage{float}
\usepackage{stfloats}
\usepackage{soul}
\usepackage{balance}
\usepackage{times}
\usepackage{graphicx} 
\usepackage{subfigure} 
\usepackage{multirow}

\newcolumntype{L}{>{\centering\arraybackslash}m{1.4cm}}
\newcolumntype{P}{>{\centering\arraybackslash}m{1.6cm}}
\newcolumntype{B}{>{\centering\arraybackslash}m{2.2cm}}
\newcolumntype{G}{>{\centering\arraybackslash}m{2.4cm}}
\newcolumntype{H}{>{\centering\arraybackslash}m{2.8cm}}
\newcolumntype{M}{>{\centering\arraybackslash}m{3.5cm}}

\begin{document}
\title{Structural Compression of Convolutional Neural Networks}

\author{Reza~Abbasi-Asl and
        Bin~Yu
\thanks{Reza Abbasi-Asl is with the Department of Neurology and Weill Institute for Neuroscience, University of California, San Francisco, CA, 94158 USA, and Department of Electrical Engineering and Computer Sciences, University of California, Berkeley, CA 94720, USA e-mail: Reza.AbbasiAsl@ucsf.edu.}
\thanks{Bin Yu is with the Departments of Statistics and Electrical Engineering and Computer Sciences, University of California, Berkeley,
CA, 94720 USA e-mail: binyu@berkeley.edu.}}

\markboth{}%
{Shell \MakeLowercase{\textit{et al.}}: Bare Demo of IEEEtran.cls for IEEE Journals}


\IEEEtitleabstractindextext{%
\begin{abstract}
Deep convolutional neural networks (CNNs) have been successful in many tasks in machine vision, however, millions of weights in the form of thousands of convolutional filters in CNNs makes them difficult for human intepretation or understanding in science. In this article, we introduce CAR, a greedy structural compression scheme to obtain smaller and more interpretable CNNs, while achieving close to original accuracy. The compression is based on pruning filters with the least contribution to the classification accuracy. We demonstrate the interpretability of CAR-compressed CNNs by showing that our algorithm prunes filters with visually redundant functionalities such as color filters. These compressed networks are easier to interpret because they retain the filter diversity of uncompressed networks with order of magnitude less filters. Finally, a variant of CAR is introduced to quantify the importance of each image category to each CNN filter. Specifically, the most and the least important class labels are shown to be meaningful interpretations of each filter.

\end{abstract}

}
\maketitle
\IEEEdisplaynontitleabstractindextext
\IEEEpeerreviewmaketitle

\section{Introduction}

\IEEEPARstart{D}{eep} convolutional neural networks (CNNs) achieve state-of-the-art performance for a wide variety of tasks in computer vision, such as image classification and segmentation \cite{krizhevsky2012imagenet, long2015fully}. Recent studies have also shown that representations extracted from these networks can shed light on new tasks through transfer learning \cite{oquab2014learning}. The superior performance of CNNs for large training datasets has led to their ubiquity in many industrial applications and to their emerging applications in science and medicine. Thus, CNNs are widely employed in many data-driven platforms such as cellphones, smart watches and robots. While the huge number of weights and convolutional filters in deep CNNs is key factor in their success, it makes them hard or impossible to interpret in general and especially for scientific and medical applications \cite{montavon2017methods, abbasi2018deeptune}. Compressing CNNs or reducing the number of weights, while keeping prediction performance, thus facilitates interpretation, and understanding in science and medicine. Moreover, compression benefits the use of CNNs in platforms with limited memory and computational power.

In this paper, interpretability is defined as the ability to explain or to present the decisions made by the model in understandable terms to a human \cite{doshi2017towards}, say a biologist or a radiologist. Interpretability is typically studied from one of two perspectives. The first is algorithmic interpretablity and transparency of the learning mechanism. The other is post-hoc interpretability and explanation of the learned model using tools such as visualization. The first perspective attempts to answer the question that how the model learns and works, while the second perspective describes the predictions without explaining the learning mechanism. From the perspective of post-hoc interpretability, a CNN with fewer filters is easier to visualize and explain to human users, because CNNs are often visualized using graphical explanation of their filters \cite{zeiler2014visualizing}. Thus to make more interpretable CNNs, a compression scheme should reduce the number of filters while keeping the model accurate (predictively). We call such schemes "structural compression". In this paper, we argue that structurally compressed networks with fewer numbers of filters are easier to be investigated or interpreted by humans for possible domain knowledge gain. 

The problem of compressing deep CNNs have been widely studied in literature, even though interpretability is not a motivating factor in majority of these studies. In the classical approach to compression of CNNs, individual weights, and not filters, are pruned and quantized \cite{han2015deep}. We call these classical compression schemes "weight compression". Optimal brain damage \cite{lecun1989optimal}, optimal brain surgeon \cite{hassibi1993optimal}, Deep Compression \cite{han2015deep}, binary neural networks \cite{rastegari2016xnor,kim1995geometrical,courbariaux2015binaryconnect} and most recently SqueezeNet \cite{iandola2016squeezenet} are some examples.

On the other hand, some studies have investigated pruning filters instead of weights, however, the interpretability of pruned networks has not been studied in details \cite{liu2017learning, he2017channel, wen2016learning, alvarez2016learning, luo2017thinet, zhao2019variational, peng2019collaborative, liu2019metapruning, you2019gate, li2019exploiting}. These studies are focused on high compression rates and low memory usage. In this paper, our goal is not to achieve state-of-the-art compression ratio or memory usage rates, but we aim to investigate the interpretability of a compressed network. However, to compare our compression ratio and computational cost to a baseline method, we chose the  structural compression in \cite{he2014reshaping, li2016pruning}. He et al. \cite{he2014reshaping} and Li et al. \cite{li2016pruning} have studied structural compression based on removing filters and introduced importance indices based on average of incoming or outgoing weights to a filter.

Pruning activations or feature-maps to achieve faster CNNs has been also studied in \cite{molchanov2016pruning}. Pruning activations can be viewed as removing filters in specific locations of the input, however, those filters almost always remain in other locations. Thus it rarely results in any compression of filters. On the other hand, pruning filters from the structure is equal to removing them for all the possible locations and avoiding to store them. Additionally, because of the simplified structure, filter-pruned networks are more interpretable compared to activation-pruned ones, therefore more applicable in scientific and medical domains.


Pruning a fully-trained neural network has a number of advantages over training the network from scratch with fewer filters. A difficulty in training a network from scratch is not knowing which architecture or how many filters to start with. While several hyper-parameter optimization techniques \cite{snoek2012practical,fernando2017pathnet,jaderberg2017population,li2017hyperband} exist, the huge numbers of possible architectures and filters would lead to a high computational cost in a combinatorial manner as in other model selection problems \cite{reed1993pruning}. Pruning provides a systematic approach to find the minimum number of filters in each layer required for accurate training. Furthermore, recent results suggest that for large-scale CNNs, the accuracy of the pruned network is slightly higher compared to a network trained from scratch (\hspace{1sp}\cite{li2017pruning} for VGG and ResNet, \cite{kim2016compression} for AlexNet). For small-scale CNNs, it is possible to train a network from scratch that achieves the same accuracy as the pruned network even though the aforementioned computational cost is not trivial in this case. Additionally, in the majority of transfer learning applications based on well-trained CNNs, pruning algorithms achieve higher accuracies compared to training from scratch given the same architecture and number of filters \cite{branson2014bird, molchanov2016pruning}. For example, \cite{branson2014bird} showed that a pruned AlexNet gains 47\% more classification accuracy in bird species categorization compared to training the network from scratch.

Our main contributions in this paper are two folds. First, we introduce a greedy structural compression scheme to prune filters in CNNs. A filter importance index is defined to be the classification accuracy reduction (CAR) (similarly for regression or RAR) of the network after pruning that filter. This is similar in spirit to the regression variable importance measures in \cite{breiman2001random, lei2018distribution}. We then iteratively prune filters in a greedy fashion based on the CAR importance index. Although achieving state of the art compression ratio is not the main goal in this paper, we show that our CAR structural compression scheme achieve higher classification accuracy in a hold-out test set compared to the baseline structural compression methods. CAR compressed AlexNet without retraining can achieve a compression ratio of 1.17 (for layer 1) to 1.5 (for layer 5) while having a close-to-original classification accuracy (54\% top-1 classification accuracy compared to original 57\%). This is 21\% (for layer 1) to 43\% (for layer 5) higher than the compression ratio from the benchmark method. If we fine-tune or retrain the CAR-compressed network, the compression ratio can be as high as 1.79 (for layer 3) when maintaining the same 54\% classification accuracy. We take advantage of weight pruning, quantization and coding by combining our method with Deep Compression \cite{han2015deep} and report considerably improved compression ratio. For AlexNet, we reduce the size of individual convolutional layers by factor of 8 (for layer 1) to 21 (for layer 3), while achieving close to original classification accuracy (or 54\% compared to 57\%) through retraining the network. 

Our second contribution is bridging the compression and interpretation for CNNs. We demonstrate the ability of our CAR algorithm to remove functionally redundant filters such as color filters making the compressed CNNs more accessible to human interpreters without much classification accuracy loss. To our knowledge, such a connection between compression and functionality has not been reported previously. Furthermore, we introduce a variant of our CAR index that quantifies the importance of each image class to each CNN filter. This variant of our CAR importance index has been presented in \cite{abbasi2017interpreting}, and is included in the Section 4 of this paper to establish the usefulness of the CAR index. Through this metric, a meaningful interpretation of each filter can be learned from the most and the least important class labels. This new interpretation of a filter is consistent with the visualized pattern selectivity of that filter.

The rest of the paper is organized as follows. In section 2, we introduce our CAR compression algorithm. The performance of the compression for the state-of-the-art CNNs in handwritten digit image and naturalistic image classification tasks is investigated in section 3.1. In section 3.2, we connect compression to the interpretation of CNNs by visualizing functionality of pruned and kept filters in a CNN. In section 4, a class-based interpretation of CNN filters using a variant of our CAR importance index is presented. The paper is concluded in section 5.

\section{CAR-based structural compression}

\subsection{Notation}

We first introduce notations. Let $w_i^L$ denote the $i^{th}$ convolutional filter in layer $L$ of the network and $n_L$ the number of filters in this layer ($i \in \{1,..,n_L\}$). Each convolutional filter is a 3-dimensional tensor with the size of $n_{L-1}\times f_L \times f_L$ where $f_L \times f_L$ is the size of spatial receptive field of the filter. 

The activation or the feature map of filter $i$ in layer $L$ ($i=1,..,n_L$) is:
$$
\alpha^L_i = f(w_i^L * \mathcal{P})
$$
where $f(\cdot)$ is the nonlinear function in convolutional network (e.g. sigmoid or ReLU) and  $\mathcal{P}$ denotes a block of activations from layer $L-1$ (i.e. the input to the neurons in layer $L$). The activation for the first layer could be patches of input images to the convolutional network. 

Assuming network $\mathcal{N}$ is trained on classification task, top-1 classification accuracy of network $\mathcal{N}$ is defined as:
$$
Acc(\mathcal{N}) = \frac{N_{Correct}}{N_{Correct}+N_{Incorrect}}
$$
where $N_{Correct}$ and $N_{Incorrect}$ are the number of correct and incorrect predicted classes, respectively. 

In this paper, we use FLOPs to quantify the computational cost in each convolutional layer of the neural network. FLOPs for each layer of the network equals to the number of floating-point operations required in that layer to classify one image. Let's assume $A \in \mathbb{R}^{n_{L-1} \times k_{L-1} \times k_{L-1}}$  is the input feature map and  $B \in \mathbb{R}^{n_{L} \times k_{L} \times k_{L}}$ is the output feature map in layer $L$ where $k_L \times k_L$ is the spatial size. The FLOPs for this convolutional layer equals to $k_{L}^2 n_{L} f_L^2 n_{L-1}$. Additionally, the storage overhead for each convolutional layer of the network equals to $4 f_L^2 n_{L-1} n_{L}$ bytes \cite{wu2016quantized}.

\subsection{The proposed algorithm based on CAR importance index}

\begin{figure}[!t]
		\begin{center}
		\centerline{\includegraphics[width=1\columnwidth]{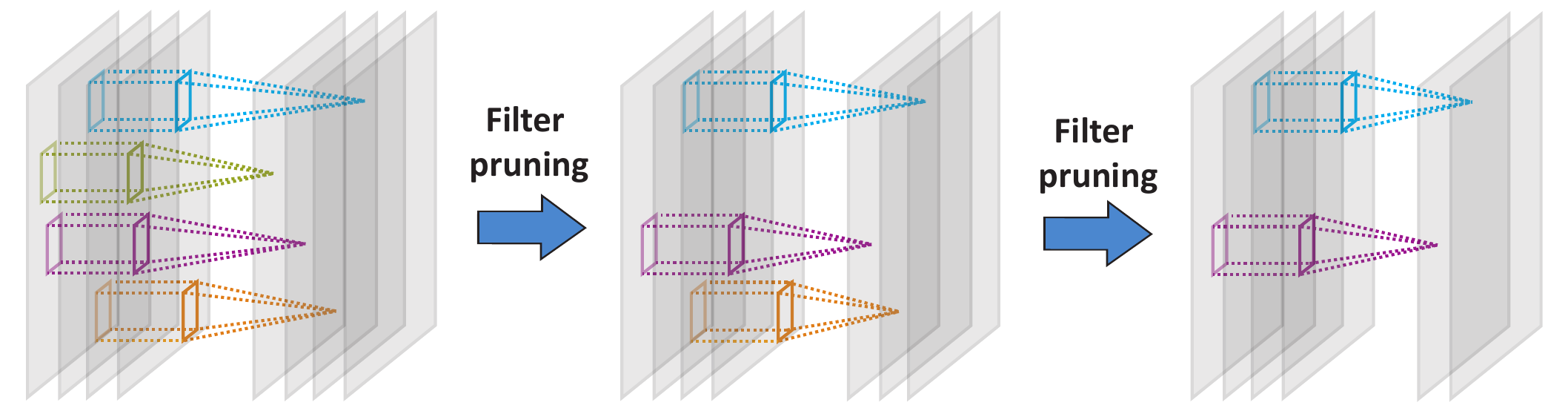}}
		\caption{Greedy compression of CNNs based on pruning filters}
		\label{compression}
		\end{center}
		\end{figure}

In this section, we introduce our greedy algorithm to prune filters in layers of a CNN and structurally compress it. Figure \ref*{compression} shows the process of greedy filter pruning. In each iteration, a candidate filter together with its connections to the next layer, gets removed from the network. The candidate filter should be selected based on an \textit{importance} index of that filter. Therefore, defining an index of importance for a filter is necessary for any structural compression algorithm. Previous works used importance indices such as average of incoming and outgoing weights to and from a filter, but with unfortunately  a considerable reduction of classification accuracy (e.g. 43\% as mentioned earlier if one prunes only the first layer)  for the compressed CNNs \cite{he2014reshaping, li2016pruning}. To overcome this limitation, we define the \textit{importance} measure for each filter in each layer as the classification accuracy reduction (CAR) when that filter is pruned from the network. This is similar in spirit to the importance measures defined for single variables in Random Forest \cite{breiman2001random} and distribution-free predictive inference \cite{lei2018distribution}.  Formally, we define CAR importance index for filter $i$ in layer $L$ of a convolutional neural network as: 
$$
CAR(i,L) =  Acc(\mathcal{N}) - Acc(\mathcal{N}(-i,L))
$$
where network $\mathcal{N}(-i,L)$ is network $\mathcal{N}$ except that filter $i$ from layer $L$ together with all of its connections to the next layer are removed from the network. In our CAR structural (or filter pruning) compression algorithm, the filter with the least effect on the classification accuracy gets pruned in each iteration. The network can be retrained in each iteration and after pruning a filter. This process is regarded as \textit{fine tuning} in this paper. We present details of our fine tuning procedure in the next section. Algorithm \ref{alg:perf} shows the pseudo code of our CAR greedy structural compression algorithm. Here, $n_{iter}$ and $r_{iter}$ are, respectively, the number of remaining filters and compression ratio in the current iteration.

		\begin{algorithm}[tb]
		   \caption{Greedy compression of CNNs based on pruning filters}
		   \label{alg:perf}
		\begin{algorithmic}
		   \STATE {\bfseries Input:} Weights in CNN, target layer $L$ with $n_L$ filters, target compression ratio $r_{target}$
		   \STATE Set $n_{iter}=n_L$ and $r_{iter} = 1$
		   \WHILE{$r_{iter} < r_{target}$}
		   \FOR{$i=1$ {\bfseries to}  $n_L$}
		   \STATE Compute $CAR(i,L)$, importance index for filter $i$ in layer $L$
		   \ENDFOR
		   \STATE Remove the least important filter, $\arg\min_i CAR(i,L)$
		   \STATE $n_{iter} = n_L-1$ 
		   \STATE Update compression rate, $r_{iter} = n_L/n_{iter}$ 
		   \ENDWHILE
		\end{algorithmic}
		\end{algorithm}

One possible drawback of the algorithm is the expensive computational cost of the early iterations. While this is a one-time computational cost for a CNN, it is still possible to significantly reduce this cost and increase the compression speed. To accomplish this, we propose the following two simple tweaks: 1. Pruning multiple filters in each iteration of the CAR algorithm. 2. Reducing the number of images for evaluating the accuracy in each iteration (i.e. batch size). Our experiments in Supplementary Materials (Figures \ref{mult_lenet1} and \ref{mult_lenet2}) suggest that the accuracy remains close to the original CAR compression, when removing multiple filters at each iteration with a smaller batch size. While these tweaks increase the compression speed, the performance of the compressed network is slightly lower than Algorithm 1. The greedy process seems to allow for a better data and network adaptation and improves compression performance. That is, when pruning one filter at each iteration, we only remove the least important filter. In the next iteration, we update all the importance indexes using the new structure. This allows the algorithm to adapt to the new structure gradually and slightly improves the classification accuracy.

The CAR compression is designed to compress each individual layer separately. This is sufficient for the majority of the transfer learning and interpretability applications, because each layer is interpreted individually. However, it is possible to compress multiple layers together too. This has been discussed in Figure \ref{multiple_layer_together} in Supplementary Materials.

\section{Results}

\subsection{Compression rate and classification accuracy of the CAR compressed networks}

To evaluate our proposed CAR structural compression algorithm, we have compressed LeNet \cite{lecun1998gradient} (with 2 convolutional layers and 20 filters in the first layer), AlexNet \cite{krizhevsky2012imagenet} (with 5 convolutional layers and 96 filters in the first layer) and ResNet-50 \cite{he2016deep} (with 50 convolutional layers and 96 filters in the first layer). LeNet is a commonly used CNN trained for classification task on MNIST \cite{lecun1998gradient} consisting of 60,000 handwritten digit images. AlexNet and ResNet-50 are trained on the subset of ImageNet dataset used in ILSVRC 2012 competition \cite{ILSVRC15} consisting of more than 1 million natural images in 1000 classes.

We used Caffe \cite{jia2014caffe} to implement our compression algorithm for CNNs and fine tune them. The pre-trained LeNet and AlexNet are obtained from Caffe model zoo. All computations were performed on an NVIDIA Tesla K80 GPU. The CAR index is computed using half of the ImageNet test set. To avoid overfitting, the final performance of CAR compressed network is evaluated on the other half of the ImageNet test set. The running time of each pruning iteration depends on the number of filters remaining in the layer. On average, each iteration of CAR takes ~45 minutes for the first layer of AlexNet. For 96 filters in this layer, the total compression time is ~72 hours. However, in Supplemental Materials, we show that it is possible prune up to five filters in one iteration without loss in the accuracy. This reduces the total running time of the compression to ~14 hours. Note that this is a one-time computational cost and much less than the time required to train AlexNet on our GPU which could take weeks.

For the fine-tuning, the learning rate has been set to 0 for the layer that is being compressed, 0.001 for the subsequent layer and 0.0001 for all other layers. The subsequent layer is directly affected by the compressed layer, therefore, requires higher learning rate. The network is retrained for 500 iterations. This is sufficent for the classification accuracy to converge to the final value. 

\subsubsection{LeNet on MNIST dataset}

\begin{figure*}[!t]
		\begin{center}
		\centerline{\includegraphics[width=1.8\columnwidth]{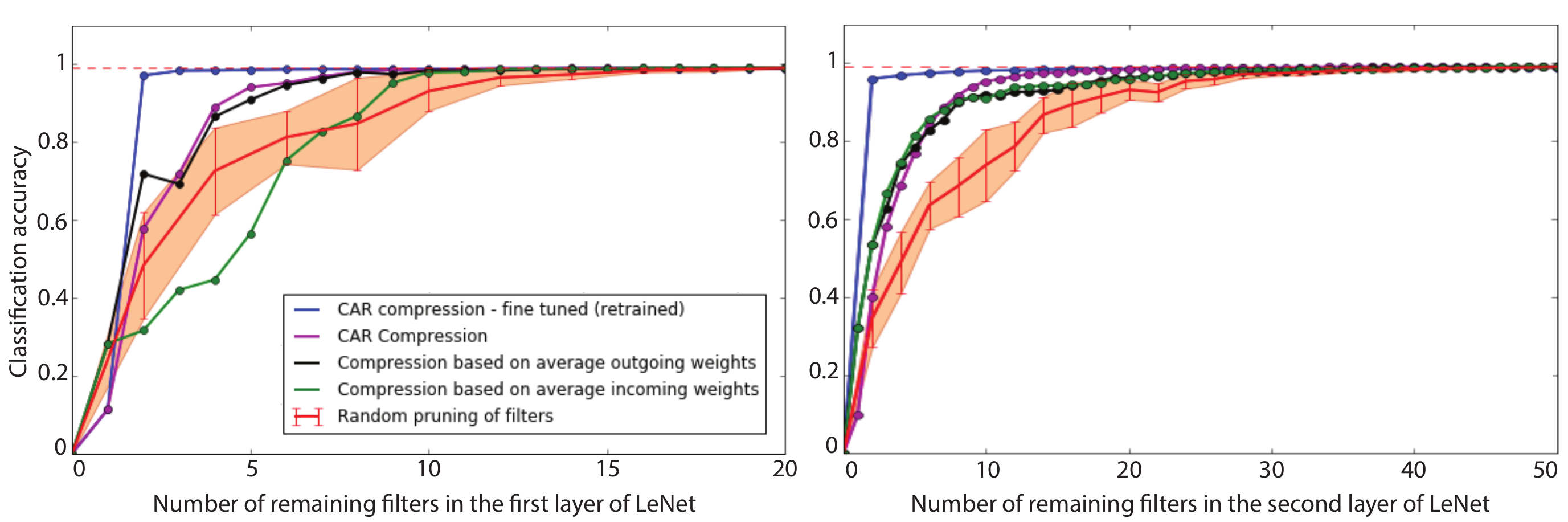}}
		\caption{Performance of compression for LeNet. The top figure shows the overall classification accuracy of LeNet when the first convolutional layer is compressed. The bottom figure shows the classification accuracy when the second convolutional layer is compressed. The classification accuracy of uncompressed network is shown with a dashed red line.  The purple curve shows the classification accuracy of our proposed CAR compression algorithm for various compression ratios.  The accuracy for the fine tuned (retrained) CAR compression is shown in blue. The black and green curves shows the accuracy for compressed network based on outgoing and incoming weights, respectively. The red curve shows the accuracy when filters are pruned uniformly at random. The error bas is reported over 10 repeats of this random pruning process. }
		\label{lenet}
		\end{center}
\end{figure*}

LeNet-5 is a four-layer CNN consisting of two convolutional layers and two fully-connected layers. CAR-compression has been performed on the convolutional layers and the performance on a hold-out test set is reported in Figure \ref{lenet}. We obtained classification accuracies (top-1) of the CAR-compression results (purple curve) and those from retraining or fine-tuning after CAR-compression on the same classification task (blue curve).

To compare the performance of our compression algorithm to benchmark filter pruning schemes, we have also implemented the compression algorithm based on pruning incoming and outgoing weights proposed in \cite{he2014reshaping} and reported the classification accuracy curve in Figure \ref{lenet}. Furthermore, classification accuracy for random pruning of filters in LeNet has been shown in this figure. Candidate filters to prune are selected uniformly at random in this case. The error bar shows the standard deviation over 10 repeats of this random selection. 

We conclude that our CAR-algorithm gives a similar classification accuracy to \cite{he2014reshaping} for LeNet  (using the outgoing weights in the first layer,  and either weights for the second layer). Their accuracies are similar to the accuracy of the uncompressed, unless we keep very few filters for either layer.  Fine-tuning improves the classification accuracy but there is not a considerable gap among performances (unless we keep very few filters, less than 8 among 20 for the first layer or less than 10 among 50 for the second layer).  Among the 8 kept filters in the first layer, 4 of them are shared between the CAR-algorithm and that based on averaging outgoing weights in \cite{he2014reshaping}, while among the 10 kept filters in the second layer, 6 of them are shared.

\subsubsection{AlexNet on ImageNet dataset}

\begin{figure*}[!t]
		\begin{center}
		\centerline{\includegraphics[width=1.8\columnwidth]{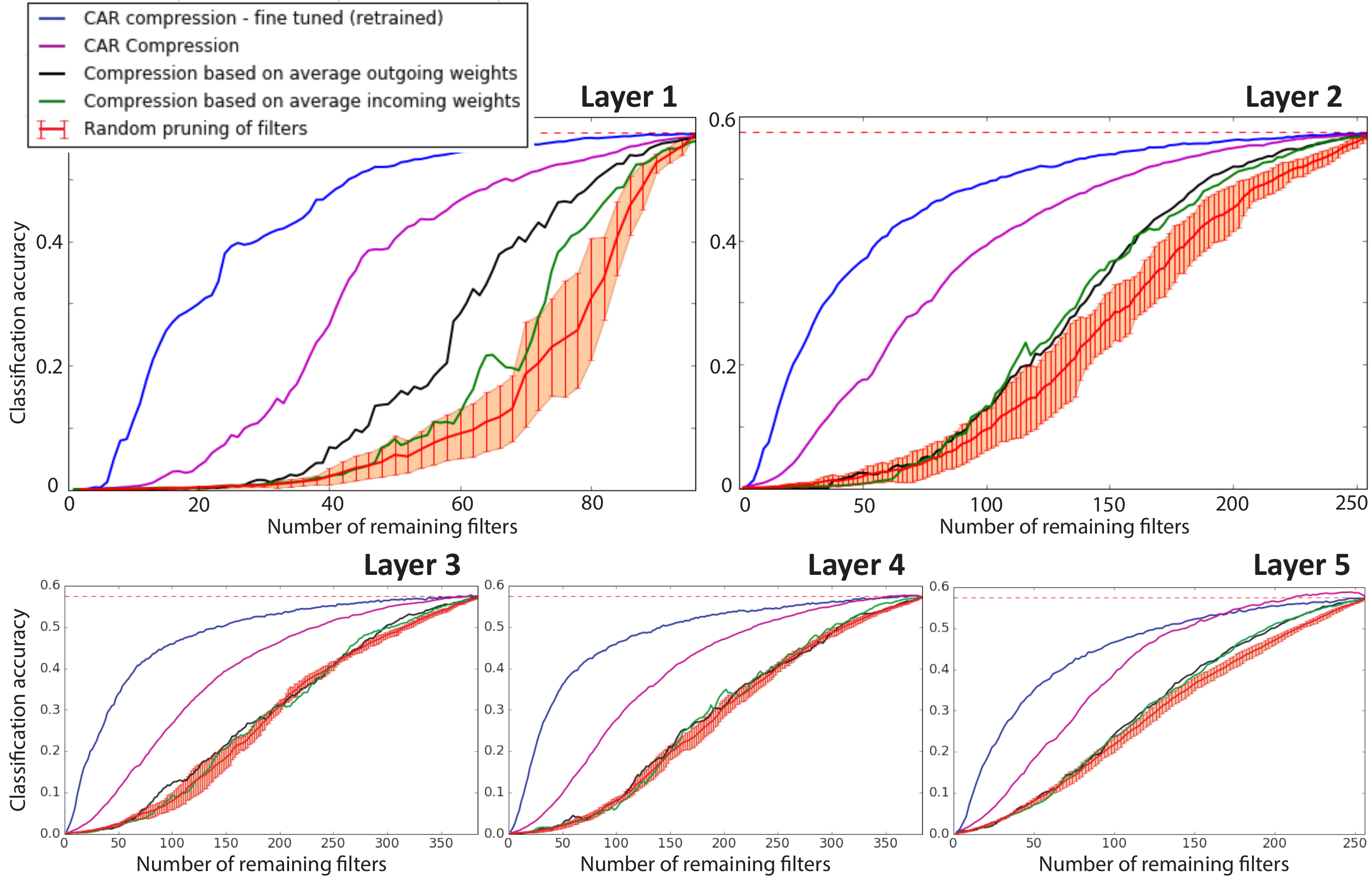}}
		\caption{Performance of compression for AlexNet. Each panel shows the classification accuracy of the AlexNet when an individual convolutional layer is compressed. In each panel, the classification accuracy of uncompressed network is shown with a dashed red line. The purple curve shows the classification accuracy of our proposed CAR compression algorithm for various compression ratios.  The accuracy for the fine tuned (retrained) CAR compression is shown in blue. The black and green curves shows the accuracy for compressed network based on outgoing and incoming weights, respectively. The red curve shows the accuracy when filters are pruned uniformly at random. The error bas is reported over 10 repeats of this random pruning process.}
		\label{alexnet}
		\end{center}
		\end{figure*}

AlexNet consists of 5 convolutional layers and 3 fully-connected layers. Figure \ref{alexnet} shows the classification accuracy of AlexNet on a hold-out test set after each individual convolutional layer is compressed using our proposed CAR algorithms or benchmark compression schemes.

Comparing the accuracies of compressed networks in Figure \ref{alexnet}, there are considerable gaps between our proposed CAR-algorithm (purple curves) and the competing structural compression schemes that prune filters \cite{he2014reshaping} for all five layers. Further considerable improvements are achieved by retraining or fine-tuning the CAR-compressed networks (see the blue curves in Figure \ref{alexnet}). 


       \begin{table*}[!t]
		\caption{Comparison of compression performance between our greedy CAR compression algorithm and benchmark schemes on AlexNet. For the compressed networks, the filters are pruned while the classification accuracy dropped a relative 5\% from the accuracy of original network (i.e. 54\% compared to 57\%). FLOPs equals to the number of floating-point operations required in each layer to classify one image.}
		\label{table1}
		\begin{center}
		\begin{tabular}{p{1.5cm} c P P P B }
		 \hline
		 Layer & Compression method & Number of remaining filters & Bytes &  FLOPs & Compression ratio \& Feed-forward speed up  \\ 
		 \hline
		 \hline
         \multirow{4}{*}{Layer 1} & Original & 96 & 0.14M & 105.41M & - \\ 
		 & Incoming weights & 90 & 0.13M & 98.82M & 1.07$\times$  \\ 
		 & Outgoing weights & 88 & 0.13M & 96.63M & 1.09$\times$  \\ 
         & CAR & \textbf{82} & \textbf{0.12M} & \textbf{90.04M} & \textbf{1.17$\times$}  \\ 
         \hline
         \multirow{4}{*}{Layer 2} & Original & 256 & 1.23M & 223.95M & - \\ 
		 & Incoming weights & 223 & 1.07M & 195.08M & 1.15$\times$  \\ 
		 & Outgoing weights & 217 & 1.04M & 189.83M & 1.18$\times$  \\ 
         & CAR & \textbf{189} & \textbf{0.91M} & \textbf{165.33M} & \textbf{1.35$\times$}  \\ 
         \hline
         \multirow{4}{*}{Layer 3} & Original & 384 & 3.54M & 149.52M & - \\ 
		 & Incoming weights & 342 & 3.15M & 133.17M & 1.12$\times$  \\ 
		 & Outgoing weights & 334 & 3.08M & 130.05M & 1.15$\times$  \\ 
         & CAR & \textbf{287} & \textbf{2.64M} & \textbf{111.75M} & \textbf{1.34$\times$}  \\ 
         \hline
         \multirow{4}{*}{Layer 4} & Original & 384 & 2.65M & 112.14M & - \\ 
		 & Incoming weights & 332 & 2.29M & 96.95M & 1.16$\times$  \\ 
		 & Outgoing weights & 346 & 2.40M & 101.04M & 1.11$\times$  \\ 
         & CAR & \textbf{279} & \textbf{1.93M} & \textbf{81.48M} & \textbf{1.38$\times$}  \\ 
         \hline
         \multirow{4}{*}{Layer 5} & Original & 256 & 1.77M & 74.76M & - \\ 
		 & Incoming weights & 220 & 1.52M & 64.25M & 1.16$\times$  \\ 
		 & Outgoing weights & 222 & 1.53M & 64.83M & 1.15$\times$  \\ 
         & CAR & \textbf{171} & \textbf{1.18M} & \textbf{49.94M} & \textbf{1.50$\times$}  \\

		 \hline 
         \hline
		 Layer 1 & \multirow{5}{*}{Fine-tuned CAR} & 58 & 0.08M & 63.69M & 1.66$\times$  \\ 
		 Layer 2 & & 153 & 0.73M & 133.84M & 1.67$\times$  \\ 
         Layer 3 & & 214 & 1.97M & 83.33M & 1.79$\times$  \\ 
		 Layer 4 & & 225 & 1.56M & 65.71M & 1.71$\times$  \\ 
         Layer 5 & & 176 & 1.22M & 51.40M & 1.45$\times$  \\ 
         
		\end{tabular}
		\end{center}
        \end{table*}

        \begin{table*}[!t]
		\caption{Compression performance of CAR-algorithm combined with Deep Compression}
		\label{table2}
		\begin{center}
		\begin{tabular}{M B H H H }
		 \hline
         \hline
		 Layer & Weight pruning + Quantization \cite{han2015deep}, $Acc = 0.57$ & Weight pruning + Quantization + Huffman Coding \cite{han2015deep}, $Acc = 0.57$ & CAR + Weight pruning + Quantization, $Acc = 0.54$  &  CAR + Weight pruning + Quantization + Huffman Coding, $Acc = 0.54$ \\ 
		 \hline
		 \hline
         Layer 1 & 3.07$\times$ & 4.87$\times$ & 5.13$\times$ & \textbf{8.13$\times$} \\  
 		 \hline
         Layer 2 & 6.90$\times$ & 10.60$\times$ & 11.52$\times$ & \textbf{17.70$\times$} \\ 
 		 \hline
         Layer 3 & 7.63$\times$ & 11.85$\times$ & 13.66$\times$ & \textbf{21.21$\times$} \\ 
 		 \hline
         Layer 4 & 7.09$\times$ & 10.98$\times$ & 12.13$\times$ & \textbf{18.77$\times$} \\ 
 		 \hline
         Layer 5 & 7.14$\times$ & 10.60$\times$ & 10.36$\times$ & \textbf{15.38$\times$} \\ 
 		 \hline
		\end{tabular}
		\end{center}
        \end{table*}

Pruning half of the filters in either of the individual convolutional layers in AlexNet, our CAR algorithm achieves 16\% (for the layer 5) to 25\% (for the layer 2)  higher classification accuracies compared to the best benchmark filter pruning scheme (pruning based on average outgoing weights). If we retrain the pruned network, it achieves 50\% to 52\% classification accuracy (compared with 57\% of the uncompressed AlexNet). The superior performance of our algorithm for AlexNet is due to the proposed importance index for the filters in CNN. This figure demonstrates that our algorithm is able to successfully identify the least important filters for the purpose of classification accuracy. In section 5.2, we discuss the ability of our compression scheme to reduce functional redundancy in the structure of CNNs.

\begin{figure*}[!t]
		\begin{center}
		\centerline{\includegraphics[width=1.7\columnwidth]{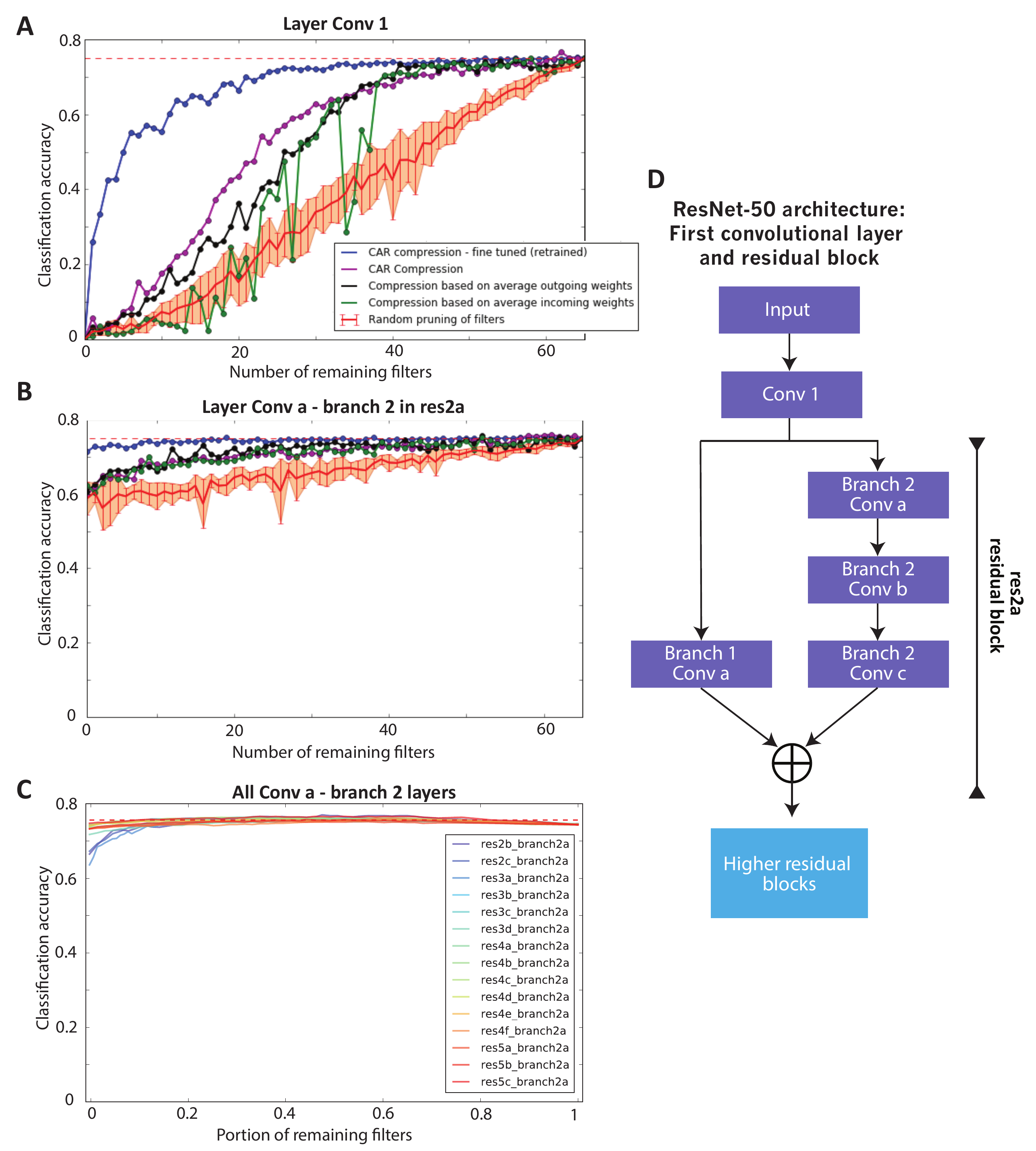}}
		\caption{Performance of compression for ResNet-50. \textbf{A.} Classification accuracy of the ResNet-50 for the compression of first convolutional layer. The classification accuracy of uncompressed network is shown with a dashed red line. The purple curve shows the classification accuracy of our proposed CAR compression algorithm for various compression ratios.  The accuracy for the fine tuned (retrained) CAR compression is shown in blue. The black and green curves shows the accuracy for compressed network based on outgoing and incoming weights, respectively. The red curve shows the accuracy when filters are pruned uniformly at random. The error bas is reported over 10 repeats of this random pruning process. \textbf{B.} Classification accuracy for the compression of first residual module (with the first layer untouched). \textbf{C.} Classification accuracy for the compression of each residual module in ResNet-50. \textbf{D.} The architecture of first layers in ResNet-50.}
		\label{resnet}
		\end{center}
\end{figure*}

To present a different but equivalent quantitative comparison, we have reported the compression ratio and feed-forward speed up in Table \ref{table1}. Each individual convolutional filter is pruned while the classification accuracy dropped a relative 5\% from the accuracy of uncompressed network (i.e. 54\% compared to 57\%). Results for CAR compression with and without fine tuning and compression based on average incoming and outgoing weights are presented in this table. The CAR algorithm (without retraining) can achieve a compression ratio of 1.17 (for layer 1) to 1.50 (for layer 5), which is 21\% to 43\% higher than those from the benchmark methods. If we fine-tune or retrain the CAR-compressed network, the compression ratio can be as high as 1.79 (for layer 3) when maintaining the same 54\% classification accuracy. 

\subsubsection{Combination with Deep Compression}

One advantage of our CAR-algorithm is that it is amenable to combination with weight based compression schemes to achieve substantial reduction in memory usage. Deep Compression \cite{han2015deep} is a recent weight-based compression procedure that uses weight pruning, quantization and huffman coding. We have performed Deep Compression on top of our proposed compression algorithm and reported the compression ratio for AlexNet in Table \ref{table2}. Again, the filters are pruned while the classification accuracy is in the range of relative 5\% from the accuracy of uncompressed network (54\% compared to 57\%). An additional 5 fold (for layer 1) to 12 fold (for layer 3) increase in compression ratio is acheived through joint CAR and Deep Compression. That is, further weight compression boosts the compression ratio by sparsifying weights of the kept filters, although the number of filters is the same as the CAR compression.


\subsubsection{ResNet-50 on ImageNet dataset}

First introduced by He et al. \cite{he2016deep}, deep residual networks take advantage of a residual block in their architecture (Figure \ref{resnet}, right panel) to achieve higher classification accuracy compared to a simple convolutional network. We have studied the performance of CAR compression on ResNet-50 architecture \cite{he2016deep} with 50 layers of convolutional weights. Figure \ref{resnet}.A shows the classification accuracy of ResNet-50 after pruning first convolutional layer using CAR algorithm or benchmark compression schemes. Figure \ref{resnet}.B shows the classification accuracy after pruning the first convolutional layer in the first residual block (layer \emph{Conv a - Branch 2} in Figure \ref{resnet}). The performance for all other residual blocks are illustrated in Figure \ref{resnet}.C. CAR pruning of other convolutional layers in each residual block yields to similar figures and are not shown here. All of the accuracies are reported on the ILSVRC 2012 ImageNet hold out test set. 

It is of great interest to compare at high compression ratio regimes where we keep less than 30 filters out of 64. In this situation and pruning layer \emph{Conv 1}, the CAR algorithm (purple curve in Figure \ref{resnet}) outperforms the competitors based on incoming and outgoing weights. The higher the compression ratio, the higher the improvements by the CAR algorithm. For low compression ratio regimes, the performances are similar. Compared to AlexNet, the gap between CAR and benchmark compressions is smaller for the first layer. This might be an evidence that ResNet has less redundant filters. Retraining (fine-tuning) the CAR-compressed network achieves further improvement in classification accuracy (blue curve in Figure \ref{resnet}). In fact, our CAR-algorithm achieves 72\% classification accuracy (compared with the 75\% for the uncompressed ResNet-50) when pruning half of the filters in the first layer of ResNet-50. This accuracy is 15\% higher than that of filter pruning based on average outgoing or incoming weights.

For the residual block, we have pruned layer \emph{Conv a - Branch 2} and reported the classification accuracy in Figure \ref{resnet}. The accuracy of CAR algorithm is almost similar to the compression based on incoming and outgoing weights. Interestingly, the accuracy drops less than 15\% if we fully prune the filters in this layer i.e. remove branch 2 from the residual block. The drop in accuracy is less than 5\% for the fine-tuned network. The main reason for this is the existence of shortcuts in the residual module. The uncompressed branch 1 that is a parallel channel with the pruned filter allows the information to transfer through the residual layer. As a result of these parallel channels in the residual blocks, deep residual networks are more robust to pruning filters compared to simple convolutional networks.

\subsection{CAR-compression algorithm prunes visually redundant filters}

\begin{figure}[t]
		\begin{center}
		\centerline{\includegraphics[width=1\columnwidth]{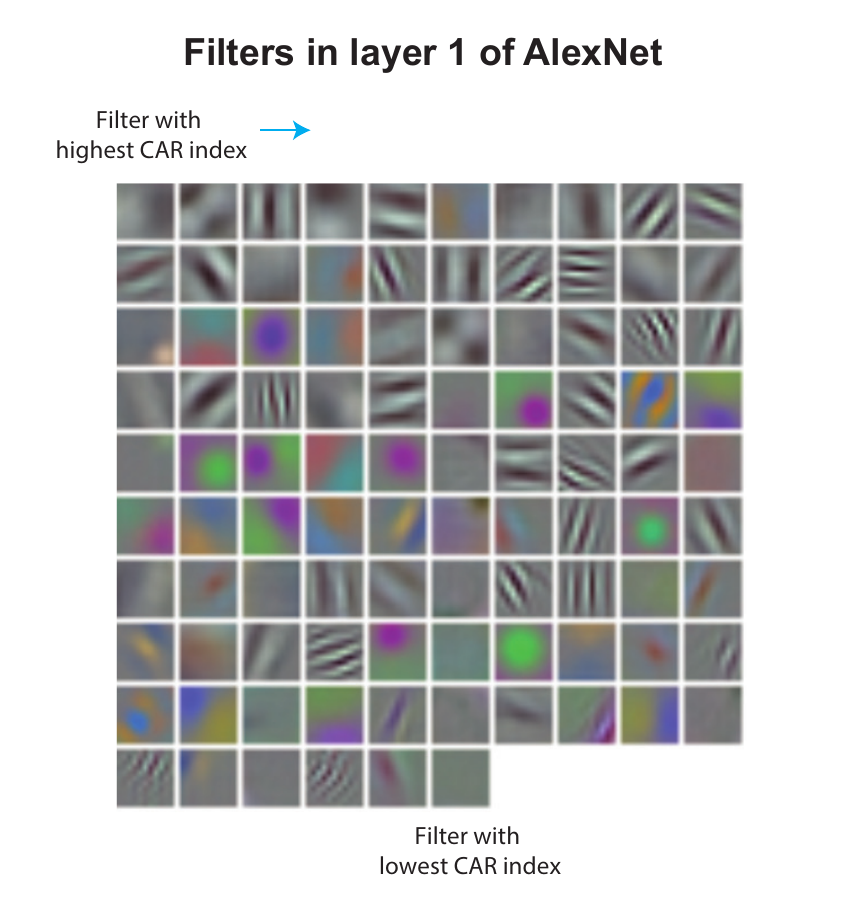}}
		\caption{Visualization of filters in the first layer of AlexNet, ordered by their CAR importance index.}
		\label{conv1}
		\end{center}
\end{figure}

To study the ability of CAR compression in identifying redundant filters in CNNs, we take a closer look at the visualization of pruned filters. Filters in the first layer of a CNN can be visualized directly using their weights (weights in the first layer filters correspond to RGB channels of the input image). Figure \ref{conv1} shows the the visualized filters in the first layer of AlexNet, ordered by their CAR importance index. Filters with higher CAR index tend to form a set of diverse patterns, spanning different orientations and spatial frequencies. Additionally, most of the filters with color selectivity tend to have lower CAR index. In fact, out of top 20 pruned filters, 15 of them in the first layer and 14 of them in the second layer correspond to the color filters, respectively. This finding points to the fact that shape is often first-order important for object recognition.

\begin{figure*}[!t]
		\begin{center}
		\centerline{\includegraphics[width=1.8\columnwidth]{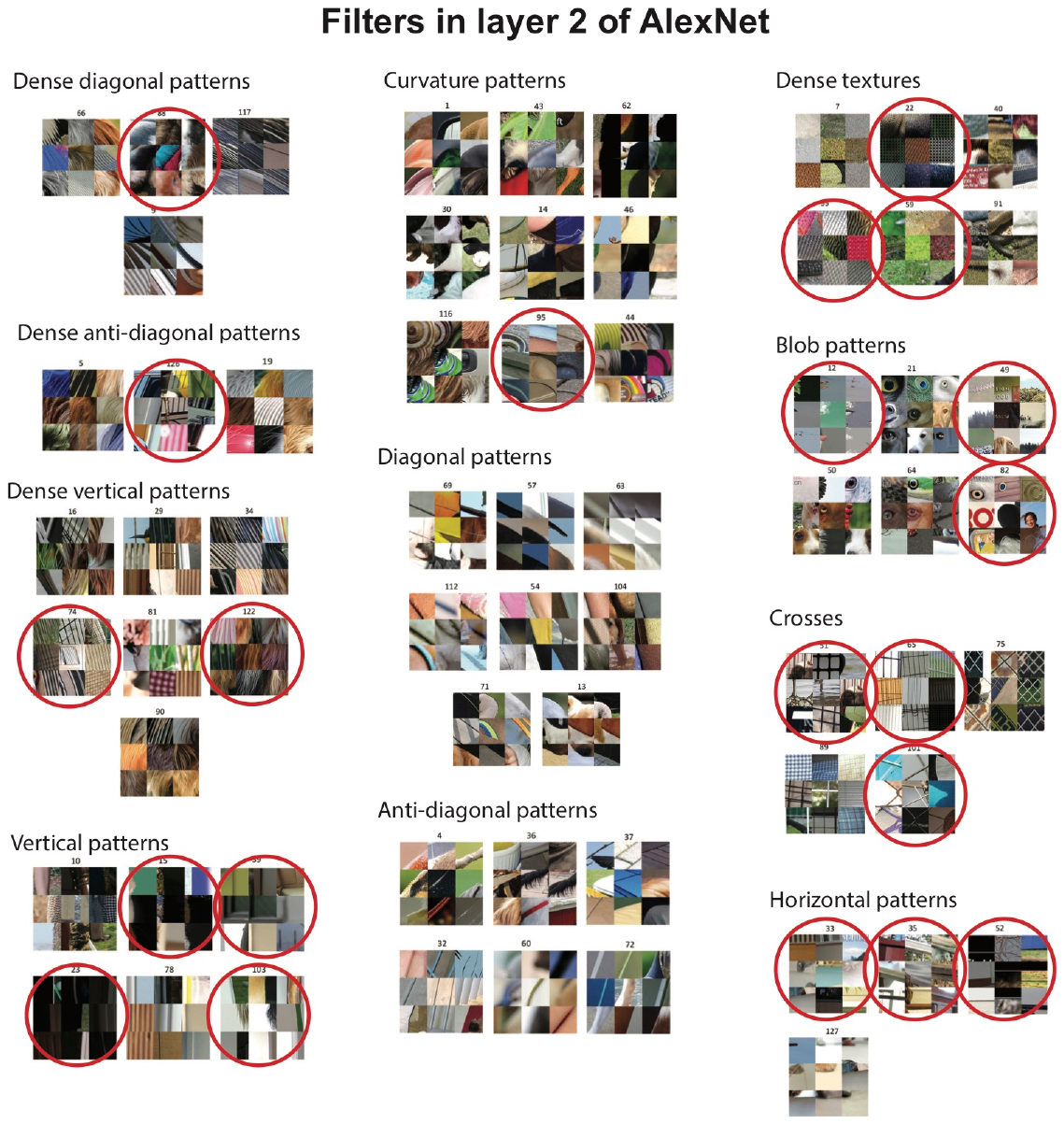}}
		\caption{CAR compression removes filters with visually redundant functionality from second layer of AlexNet. To visualize each filter, we have fed one million image patch to the network and visualized each filter by 9 image patches with top response for that filter. We have manually clustered 256 filters in the second layer of AlexNet into 20 groups (11 of them visualized here) based on their pattern selectivity. We continue to iterate the CAR-based algorithm while the classification accuracy is in the relative range of 5\% from the accuracy of uncompressed network. This leads to pruning 103 out of 256 filters in this layer. The pruned filters are specified with a red circle. }
		\label{redundent}
		\end{center}
\end{figure*}

Unlike the first layer, visualization of filters in higher convolutional layers is not trivial. To visualize the pattern selectivity of filters in these higher layers, we have fed one million image patch to the network and showed the top 9 image patch that activate each filter. This approach has been previously used to study functionality of filters in deep CNNs \cite{zeiler2014visualizing}. There are 256 filters in the layer 2 of AlexNet which makes it challenging to visualize all of these filters. Therefore, we manually grouped filters into subsets with visually similar pattern selectivity in Figure\ref{redundent}. To investigate the ability of CAR compression in removing visually redundant filters in this layer, we continued to iterate the CAR algorithm while the classification accuracy is 54\% or within a relative 5\% from the accuracy of uncompressed network. This led to pruning 103 filters out of 256 filters in the second layer. A subset of the removed and remaining filters are visualized in Figure \ref{redundent}. The filters shown with red circles are pruned in the CAR process. Similar figures for other layers are shown in Supplementary Materials (Figures \ref{conv3} and \ref{conv4}). Our algorithm tends to keep at least one filter from each group, suggesting that our greedy filter pruning process is able to identify redundant filters. This indicates that pruned filters based on the CAR importance index have in fact redundant functionality in the network. 

\begin{figure*}[!t]
		\begin{center}
		\centerline{\includegraphics[width=1.8\columnwidth]{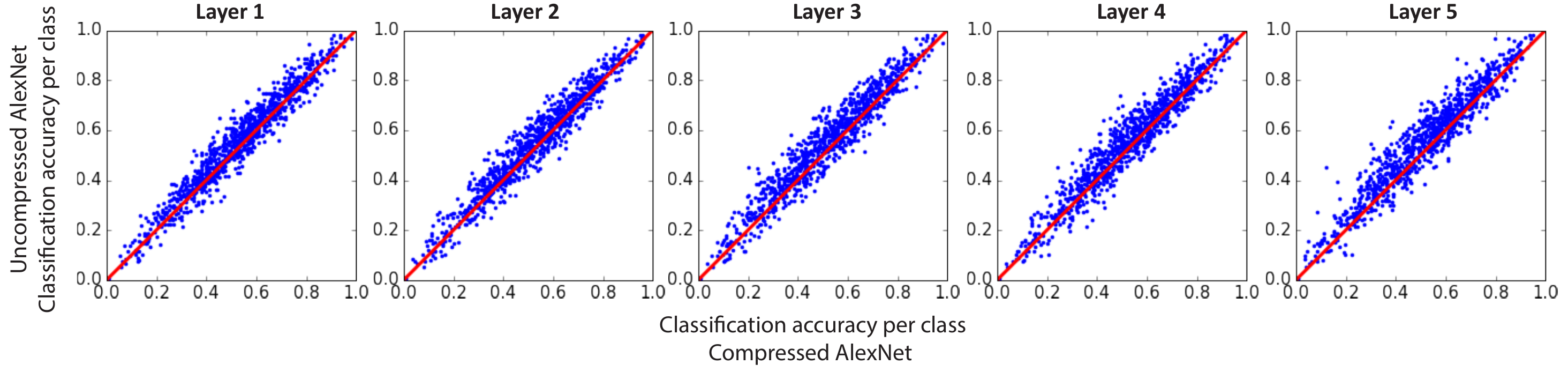}}
		\caption{Classification accuracy for each class of image in AlexNet after the first (left panel) or second layer (right panel) is compressed compared to the uncompressed network. Each point in plots corresponds to one of the 1000 categories of images in test set.}
		\label{perclass}
		\end{center}
	\end{figure*}

To further investigate the effect of compression of each of the convolutional layers, we have shown the scatter plots of the classification accuracy for each of the 1000 classes in ImageNet in Figure \ref{perclass}. Although the total classification accuracy is about a relative 5\% lower for the each compressed network, the accuracies for many of the categories are comparable between compressed and uncompressed networks. In fact, 37\% (for layer 5) to 49\% (for layer 2) of the categories have accuracies no larger than 3\% below those for the uncompressed network.


\section{Class-based interpretation of filters}

With a slight modification in the definition for the CAR importance index, we build a new index to interpret the filters via image class labels. This index has been introduced in \cite{abbasi2017interpreting} and also included in this section to demonstrate the merit of the CAR index. We define $CAR^c(i,L)$ to be classification accuracy reduction in class $c$  of images when filter $i$ in layer $L$ is pruned. $CAR^c$ quantifies the importance of each filter in predicting a class label. Therefore, for each filter, we can use $CAR^c$ to identify classes that are highly dependent on that filter (classification accuracy for these classes depends on the existence of that filter). These classes are the ones with highest $CAR^c$ among all other classes. Similarly, for each filter, the performance in classes with the lowest $CAR^c$ have less dependency to that filter. 

\begin{figure*}[!t]
		\begin{center}
		\centerline{\includegraphics[width=1.8\columnwidth]{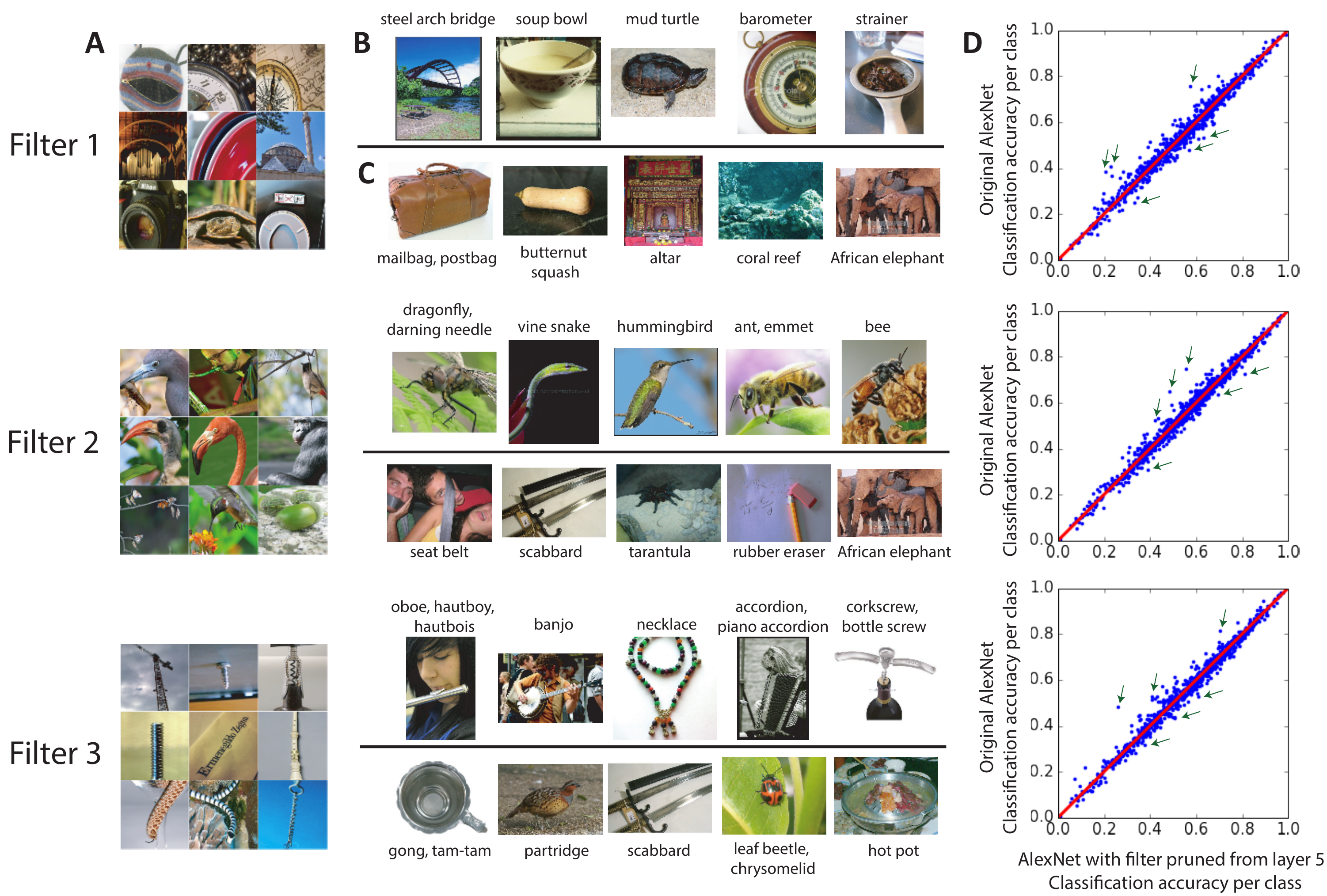}}
		\caption{The interpretation based on $CAR^c$ is consistent with the visualized pattern selectivity of each filter in layer 5 of AlexNet. Panel A shows The top 9 image patches that activate each filter \cite{zeiler2014visualizing}. Panel B and C show the top and bottom 5 classes with highest and lowest $CAR^c$, respectively. Besides the class label, one sample image from that class is also visualized. Panel D shows the scatter plot of classification accuracy for each of the 1000 classes in ImageNet. Three of the top and bottom classes with highest and lowest $CAR^c$ are pointed out with green arrows. Each row corresponds to one filter in layer 5 of AlexNet.}
		\label{perclass_interp}
		\end{center}
\end{figure*}

The labels of the two sets of classes with highest and lowest $CAR^c$ present a verbal interpretation of each filter in the network. This is particularly important in application of CNNs in scientific domains such as medicine, where it is necessary to provide a verbal explanation of the filters for the user. $CAR^c$-based interpretation is a better fit for the higher layers in the CNN because filters in these layers are more semantic and therefore more explainable by the class labels. For these layers, the interpretation of filters based on $CAR^c$ is consistent with the visualization of pattern selectivity for that filter. Figure \ref{perclass_interp} illustrates this consistency for layer 5 of AlexNet. We focus on three filters in layer 5 that are among the most important filters in this layer based on our original CAR pruning. Similar to Figure \ref{perclass}, the pattern selectivity of each filter is visualized in panel A using top 9 image patch activating that filter. Panels B and C show the top and bottom 5 classes with highest and lowest $CAR^c$, respectively. Beside the class label, one sample image from that class is also visualized. Some of these classes are pointed out with green arrows in the scatter plot of classification accuracy for 1000 classes in ImageNet (panel D). Note that both CAR and $CAR^c$ indexes could be negative numbers, that is the pruned network has higher classification accuracy compared to the original network.

In Figure \ref{perclass_interp}, the classes with highest $CAR^c$ share similar patterns with the top 9 patches activating each filter. For filter 1, the smooth elliptic curvature that consistently appears in the classes such as \textit{steep arch bridge} or \textit{soup bowel} is visible in the top activating patches (Panel A). On the other hand, less elliptic curvature patterns are expected in classes such as \textit{mailbag} or \textit{altar}. Filter 2 has higher $CAR^c$ for classes that contains patterns such as insect or bird's head. Filter 3 is mostly selected by the classes that contain images of a single long tool, particularly musical instruments such as \textit{oboe} or \textit{banjo}.

\section{Discussion and future work}

Structural compression (or filter pruning) of CNNs has the dual purposes of saving memory cost and computational cost on small devices, and of resulted CNNs being more humanly interpretable in general and for scientific and medical applications in particular. In this paper, we proposed a greedy filter pruning based on the importance index of classification accuracy reduction (CAR). We have shown with AlexNet that the huge gain (8 to 21 folds) in compression ratio of CAR+Deep Compression schemes, without a serious loss of classification accuracy. Furthermore, we saw that the pruned filters have redundant functionality for the AlexNet. In particular, for many categories in ImageNet, we found that the redundant filters are color-based instead of shape-based. This suggests the first order importance of shape for such categories.

However, a greedy algorithm is likely to be sub-optimal in identifying the best candidate filters to drop. The optimal solution may be to search through all possible subsets of filters to prune, but this can be computationally expensive and may lead to over-pruning. Procedures for subset selection, including genetic algorithms and particle swarm optimization, could be helpful in the compression of CNNs and will be investigated in future work. Even though the CAR compression of ResNet achieves state-of-the-art classification accuracy among other structural compressions by pruning the identity branch and identifying the redundant connections.  ResNet compression merits further investigation because of the identity branches in the residual blocks.

We also proposed a variant of CAR index to compare classification accuracies of original and pruned CNNs for each image class. In general, we can compare any two convolutional neural networks that are trained on the similar dataset through this index. The comparison could be done by looking into set of classes that are important for each filter in layer of each network. A similar class-based comparison for any two networks through our importance index is possible. This is a fruitful direction to pursue, particularly given the recent wave of various CNNs with different structures. Finally, we expect that our CAR structural compression algorithm for CNNs and related interpretations can be adapted to fully-connected networks with modifications. 

\section*{Acknowledgements}
This work is supported in part by National Science Foundation (NSF) Grants DMS-1613002, and IIS-1741340, and the Center for Science of Information, an NSF Science and Technology Center, under Grant Agreement CCF-0939370.

\bibliographystyle{unsrt}

\balance
\pagebreak

\pagenumbering{arabic}
\setcounter{page}{1}

\begin{center}
\textbf{\large Supplementary Materials}
\end{center}

\setcounter{section}{0}

\renewcommand\thefigure{S.\arabic{figure}}  
\setcounter{figure}{0}

\balance

\section{Boosting the compression speed}

\begin{figure*}[h]
		\begin{center}
		\centerline{\includegraphics[width=2\columnwidth]{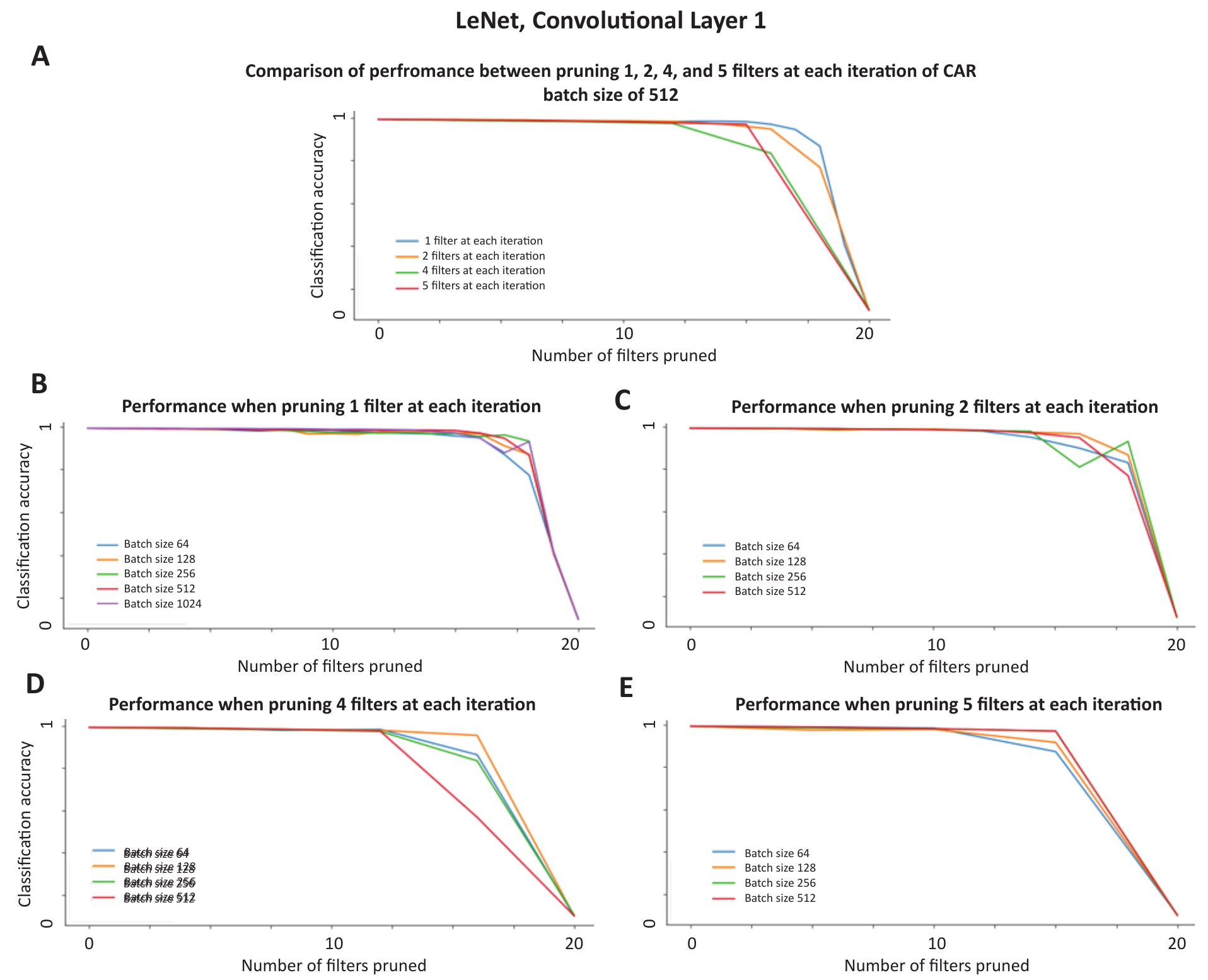}}
		\caption{Removing multiple filters at each iteration of CAR algorithm boosts compression speed without degrading the accuracy for LeNet layer 1. \textbf{A} Classification accuracy as a function of number of filters pruned in LeNet layer 1 when removing 1, 2, 4, or 5 filters at each iteration. Batch size is set to 512. The accuracy remains close to original CAR compression when pruning 5 filters at each iteration. \textbf{B} Comparison of the accuracy curves between different batch sizes when removing 1 filter at each iteration of the algorithm. \textbf{C, D, and E} Similar to B but when removing 2, 4, or 5 filters at each iteration of the algorithm, respectively. Batch size does not have a considerable effect on the curves.}
		\label{mult_lenet1}
		\end{center}
\end{figure*}

\begin{figure*}[h]
		\begin{center}
		\centerline{\includegraphics[width=2\columnwidth]{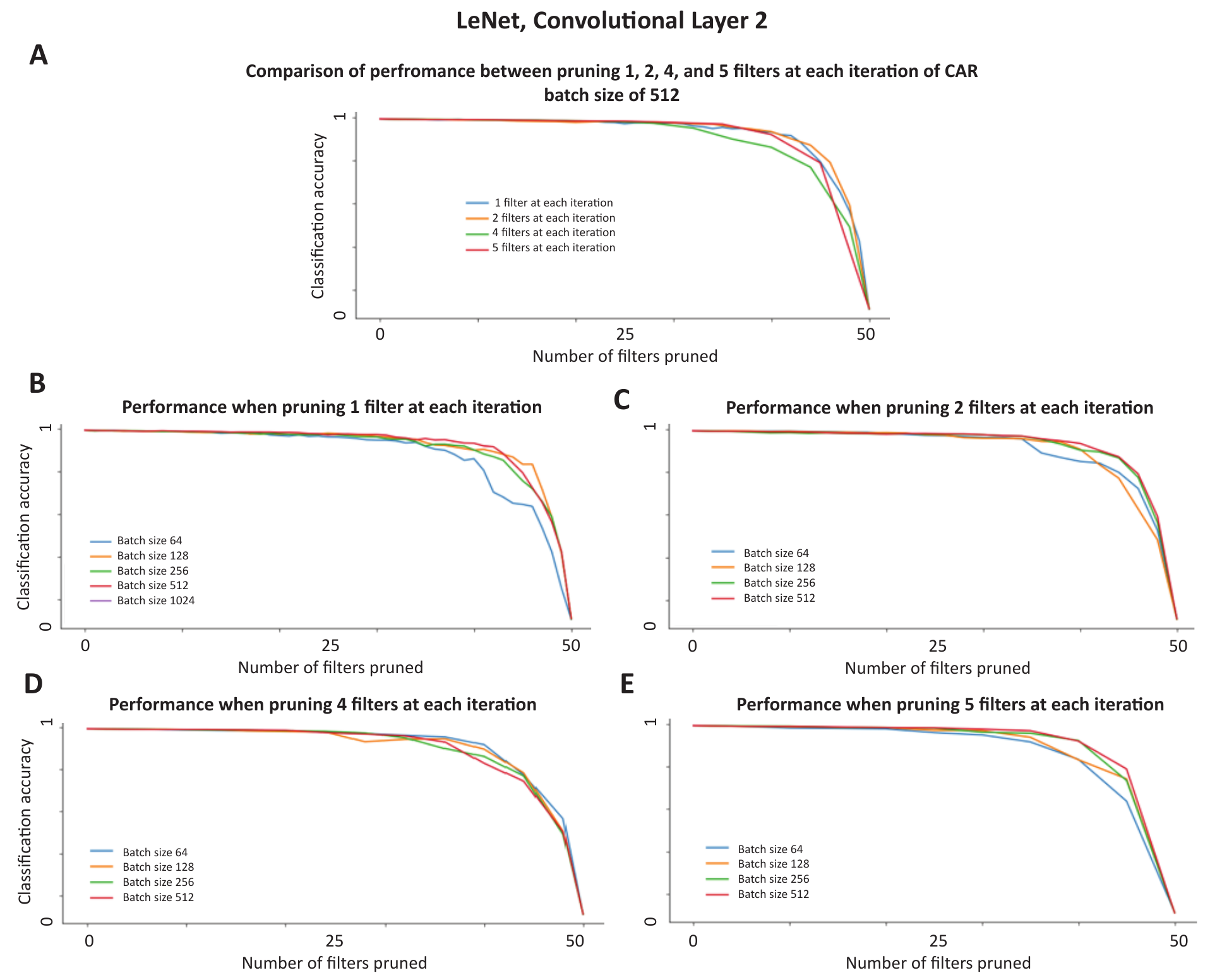}}
		\caption{Removing multiple filters at each iteration of CAR algorithm boosts compression speed without degrading the accuracy for LeNet layer 2. \textbf{A} Classification accuracy as a function of number of filters pruned in LeNet layer 2 when removing 1, 2, 4, or 5 filters at each iteration. Batch size is set to 512. The accuracy remains close to original CAR compression when pruning 5 filters at each iteration. \textbf{B} Comparison of the accuracy curves between different batch sizes when removing 1 filter at each iteration of the algorithm. \textbf{C, D, and E} Similar to B but when removing 2, 4, or 5 filters at each iteration of the algorithm, respectively. Batch size of 64 degrades the accuracy. }
		\label{mult_lenet2}
		\end{center}
\end{figure*}

While CAR compression constructs a more interpretable network with sufficiently high classification performance, the main drawback is the expensive computational cost. Here, we propose two following tweaks to increase the compression speed: 1. Pruning multiple filters in each iteration of the CAR algorithm. 2. Reducing the number of images for evaluating the accuracy in each iteration (i.e. batch size). Our experiments suggest that the accuracy remains close to the original CAR compression, when removing up to 5 filters at each iteration with a batch size of 128 for LeNet. Removing 5 filters at each iteration increased the computational speed by a factor of 5. Using the batch size of 128 instead of the 5000 increased the speed by a factor of 12. In total, the compression speed increased by a factor of 60 for LeNet while keeping the accuracy close to original CAR compression. Figures \ref{mult_lenet1}.A illustrates the classification accuracy as a function of number of filters pruned in LeNet layer 1 when removing 1, 2, 4, or 5 filters at each iteration. The accuracy remains close to original CAR compression when pruning 5 filters at each iteration. Panels B, C, D, and E in Figure \ref{mult_lenet1} compares the accuracy curves between different batch sizes. Batch size in this figure equals to the number of images from the validation set used in each iteration of CAR. For LeNet layer 1, batch size does not have a considerable effect on the curve when pruning up to 5 filters at each iteration of the algorithm. Figure \ref{mult_lenet2} illustrates the accuracy curves for LeNet layer 2. Similar to layer 1, the accuracy remains close to original CAR compression when pruning up to 5 filters at each iteration with batch size 512. However, for LeNet Layer 2, batch size of 64 degrades the accuracy.

\section{Compressing multiple layers}

\begin{figure*}[!t]
		\begin{center}
		\centerline{\includegraphics[width=1.1\columnwidth]{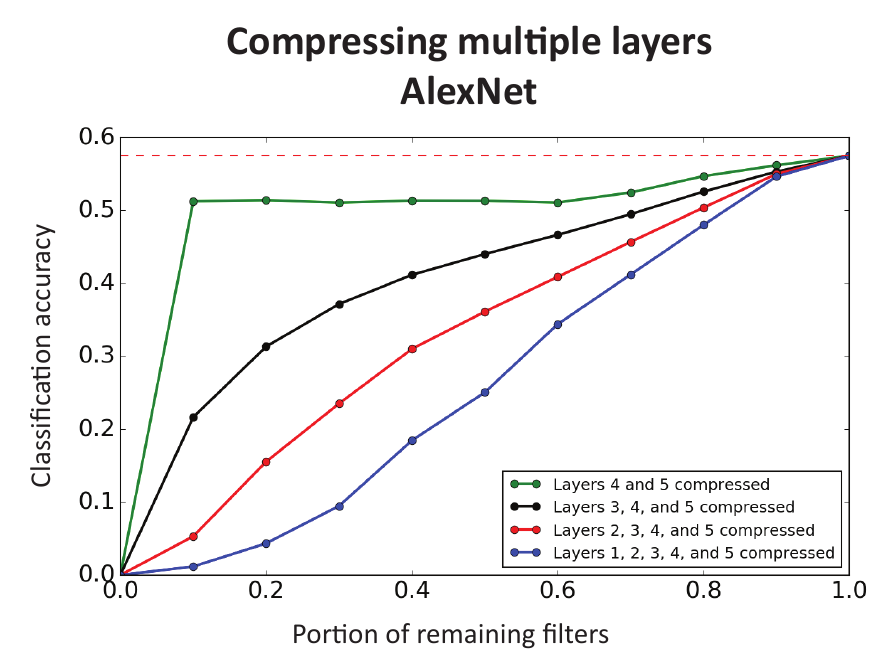}}
		\caption{Classification accuracy when compressing multiple layers together. Classification accuracy curves as a function of portion of remaining filters in the network are shown for compressing multiple combination of layers. The classification accuracy decreases as we increase the number of layers involved in the compression. }
		\label{multiple_layer_together}
		\end{center}
\end{figure*}

To study the accuracy of CAR when multiple layers are compressed together, we used CAR to compre   Figure \ref{multiple_layer_together} shows the classification accuracy curves as a function of portion of remaining filters in the network. As expected, the classification accuracy decreases as we increase the number of layers involved in the compression.

\section{CAR pruned filters for layers 3 and 4 of AlexNet}
\begin{figure*}[!t]
		\begin{center}
		\centerline{\includegraphics[width=1.47\columnwidth]{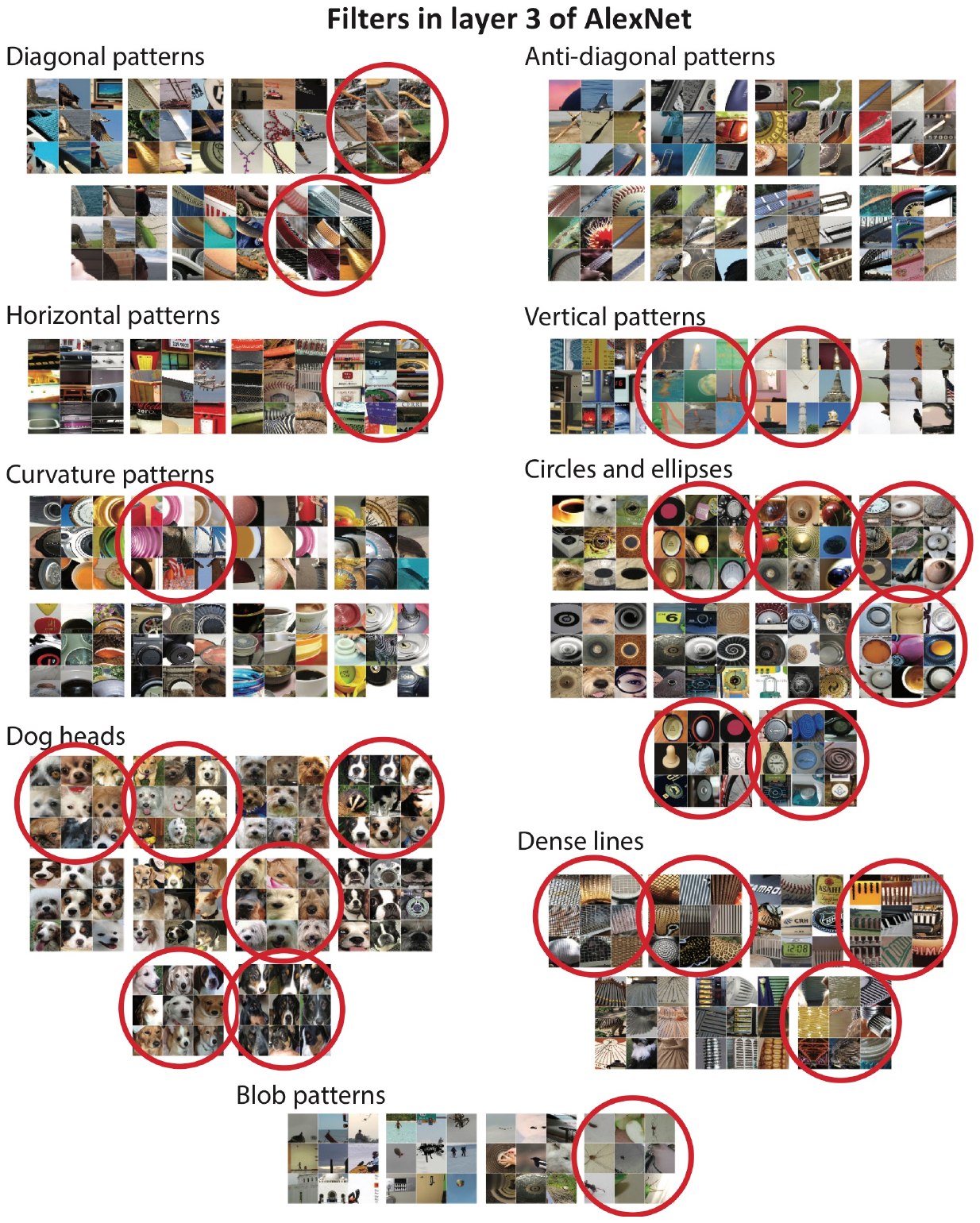}}
		\caption{CAR compression removes filters with visually redundant functionality from third layer of AlexNet. To visualize each filter, we have fed one million image patch to the network and visualized each filter by 9 image patches with top response for that filter. We have manually clustered filters in the third layer of AlexNet based on their pattern selectivity. We continue to iterate the CAR-based algorithm while the classification accuracy is in the relative range of 5\% from the accuracy of uncompressed network. This leads to pruning 170 out of 384 filters in this layer. The pruned filters are specified with a red circle. }
		\label{conv3}
		\end{center}
\end{figure*}

\begin{figure*}[!t]
		\begin{center}
		\centerline{\includegraphics[width=1.65\columnwidth]{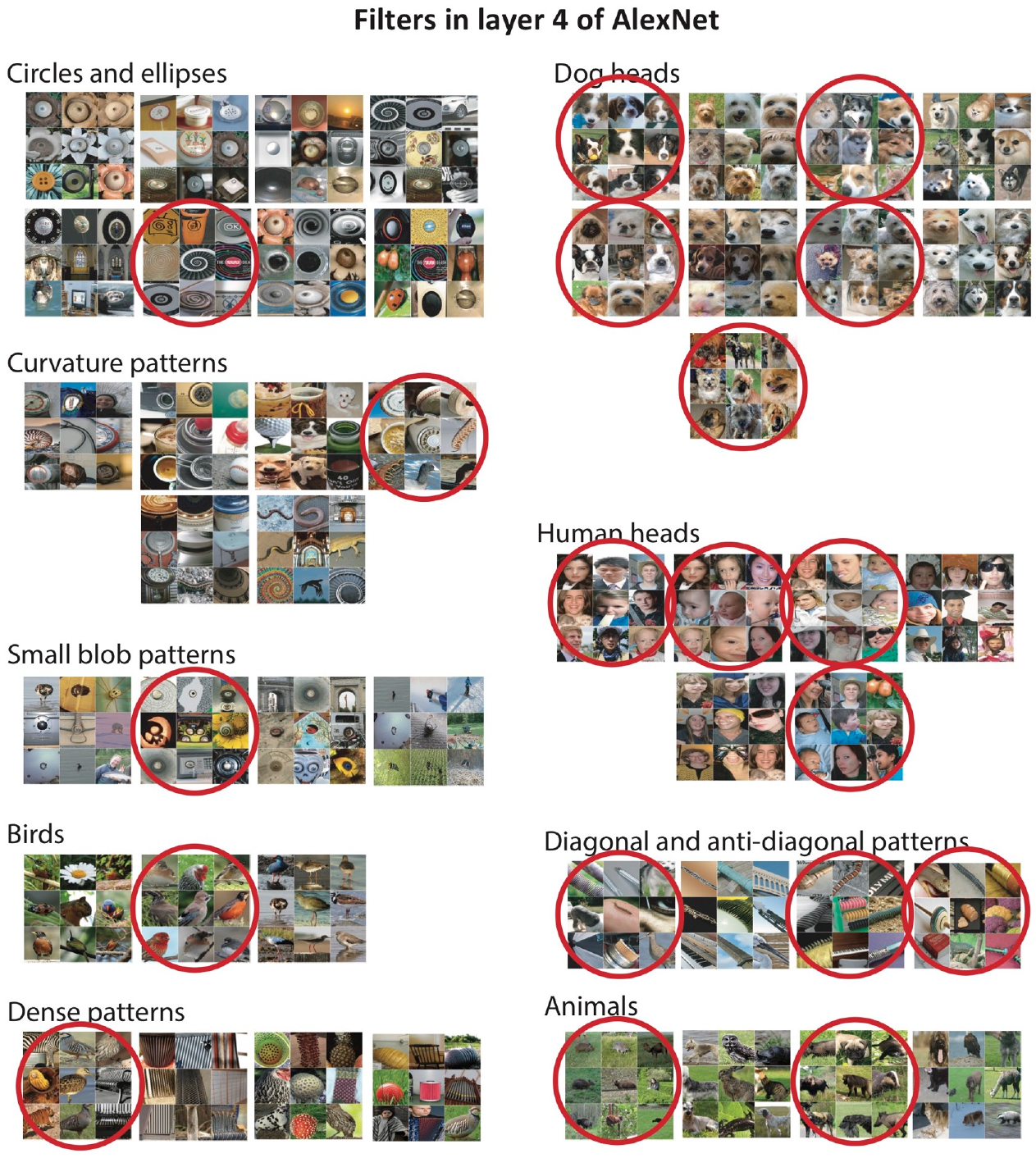}}
		\caption{CAR compression removes filters with visually redundant functionality from fourth layer of AlexNet. To visualize each filter, we have fed one million image patch to the network and visualized each filter by 9 image patches with top response for that filter. We have manually clustered filters in the third layer of AlexNet based on their pattern selectivity. We continue to iterate the CAR-based algorithm while the classification accuracy is in the relative range of 5\% from the accuracy of uncompressed network. This leads to pruning 159 out of 384 filters in this layer. The pruned filters are specified with a red circle. }
		\label{conv4}
		\end{center}
\end{figure*}

To further elaborate on the ability of CAR compression in identifying redundant filters, we have visualized filters in layers 3 and 4 of AlexNet in Figures \ref{conv3} and \ref{conv4}, respectively. Similar to Figure \ref{redundent} (in the main text) for layer 2, the filters are manually clustered based on their pattern selectivity. We continue to iterate the CAR pruning while the classification accuracy is in the relative range of 5\% from the accuracy of uncompressed network. The pruned filters are specified with a red circle. Similar to layer 2, the CAR algorithm tend to keep filters with diverse functionality in the network.

\end{document}